\documentclass[letterpaper, 10 pt, conference]{ieeeconf}  

\IEEEoverridecommandlockouts                              

\usepackage{graphicx} 
\usepackage{amsmath}  
\usepackage{amssymb}  
\usepackage{hyperref}
\usepackage{subcaption} 
\usepackage{cite}

\title{\LARGE \bf
Integrated Planning and Control on Manifolds: Factor Graph Representation and Toolkit}

\author{Peiwen Yang, Weisong Wen\thanks{*Corresponding author: Weisong Wen. Email: welson.wen@polyu.edu.hk}, Runqiu Yang, Yuanyuan Zhang, Jiahao Hu, Yingming Chen, \\
Naigui Xiao and Jiaqi Zhao
}

\begin{document}

\maketitle
\thispagestyle{empty}
\pagestyle{empty}

\begin{abstract}
Model predictive control (MPC) faces significant limitations when applied to systems evolving on nonlinear manifolds, such as robotic attitude dynamics and constrained motion planning, where traditional Euclidean formulations struggle with singularities, over-parameterization, and poor convergence. To overcome these challenges, this paper introduces FactorMPC, a factor-graph based MPC toolkit that unifies system dynamics, constraints, and objectives into a modular, user-friendly, and efficient optimization structure. Our approach natively supports manifold-valued states with Gaussian uncertainties modeled in tangent spaces. By exploiting the sparsity and probabilistic structure of factor graphs, the toolkit achieves real-time performance even for high-dimensional systems with complex constraints. The velocity-extended on-manifold control barrier function (CBF)-based obstacle avoidance factors are designed for safety-critical applications. By bridging graphical models with safety-critical MPC, our work offers a scalable and geometrically consistent framework for integrated planning and control. The simulations and experimental results on the quadrotor demonstrate superior trajectory tracking and obstacle avoidance performance compared to baseline methods. To foster research reproducibility, we have provided open-source implementation offering plug-and-play factors. Code and supplementary materials available at: \url{https://anonymous.4open.science/r/FactorMPC-EBF0}.
\end{abstract}

\section{Introduction}
Model predictive control (MPC) has become a fundamental methodology for optimizing the behavior of dynamic systems \cite{ref1,ref2,ref3} in robotics, such as multirotor aerial vehicles (MAVs). However, traditional MPC approaches often encounter challenges when applied to systems governed by nonlinear dynamics or evolving on non-linear manifolds, such as $SO(3)$ for spacecraft attitude control \cite{ref4} and constrained motion planning \cite{ref5,ref6}, where Euclidean assumptions break down, leading to singularities and poor convergence. Therefore, we propose a novel integrated planning and control framework that uses factor graph representations \cite{ref7,ref8,ref9,ref10} to enhance flexibility and computational efficiency.

In real-world deployments, robotic systems must operate reliably under dynamic uncertainties and stringent safety constraints. Robust \cite{ref11} and stochastic \cite{ref12,ref13,ref14} MPC formulations account for bounded or probabilistic uncertainties, while control barrier functions (CBFs) \cite{ref15,ref16,ref17} provide a theoretical framework for ensuring forward invariance of safe sets \cite{ref17}, making them ideal for integration with optimization-based controllers \cite{ref15,ref17}. Additionally, MPC's explicit constraint handling inspires neural networks to learn safety mechanisms. For example, a neural graph CBF framework was proposed for distributed safe multi-agent control \cite{ref18}.

The open-source solvers ACADO \cite{ref19}, CasADi \cite{ref20}, OSQP \cite{ref21}, and qpOASES \cite{ref22} are commonly used in MPC-related research. While powerful, the effective use of these solvers hinges on the ability to formulate the underlying optimization problem---be it a quadratic programming (QP) or nonlinear programming (NLP)---in a way that is both efficient and numerically stable. This is particularly challenging for systems on manifolds, where traditional Euclidean parameterizations can lead to ill-conditioning. To overcome this, we leverage factor graphs \cite{ref7,ref23}, which provide a unified and probabilistic representation for intuitively encoding system dynamics, constraints, and objectives in a modular fashion. This approach directly addresses the challenges of solver integration: the inherent sparsity of factor graphs supports efficient inference \cite{ref23,ref24} for large problems (complementing solvers like OSQP), while their native compatibility with manifold geometry \cite{ref25} ensures numerical stability and accurate state representation---bypassing the singularities that complicate the use of generic NLP tools like CasADi. Crucially, this framework abstracts complexity; cost functions and constraints are encapsulated as simple 'factors', allowing users to easily modify and assemble complex problems without deep solver expertise. The convergence of each component can also be intuitively monitored through individual factor residuals, providing valuable debugging insight.
\begin{figure*}[thpb]
\centering
\includegraphics[]{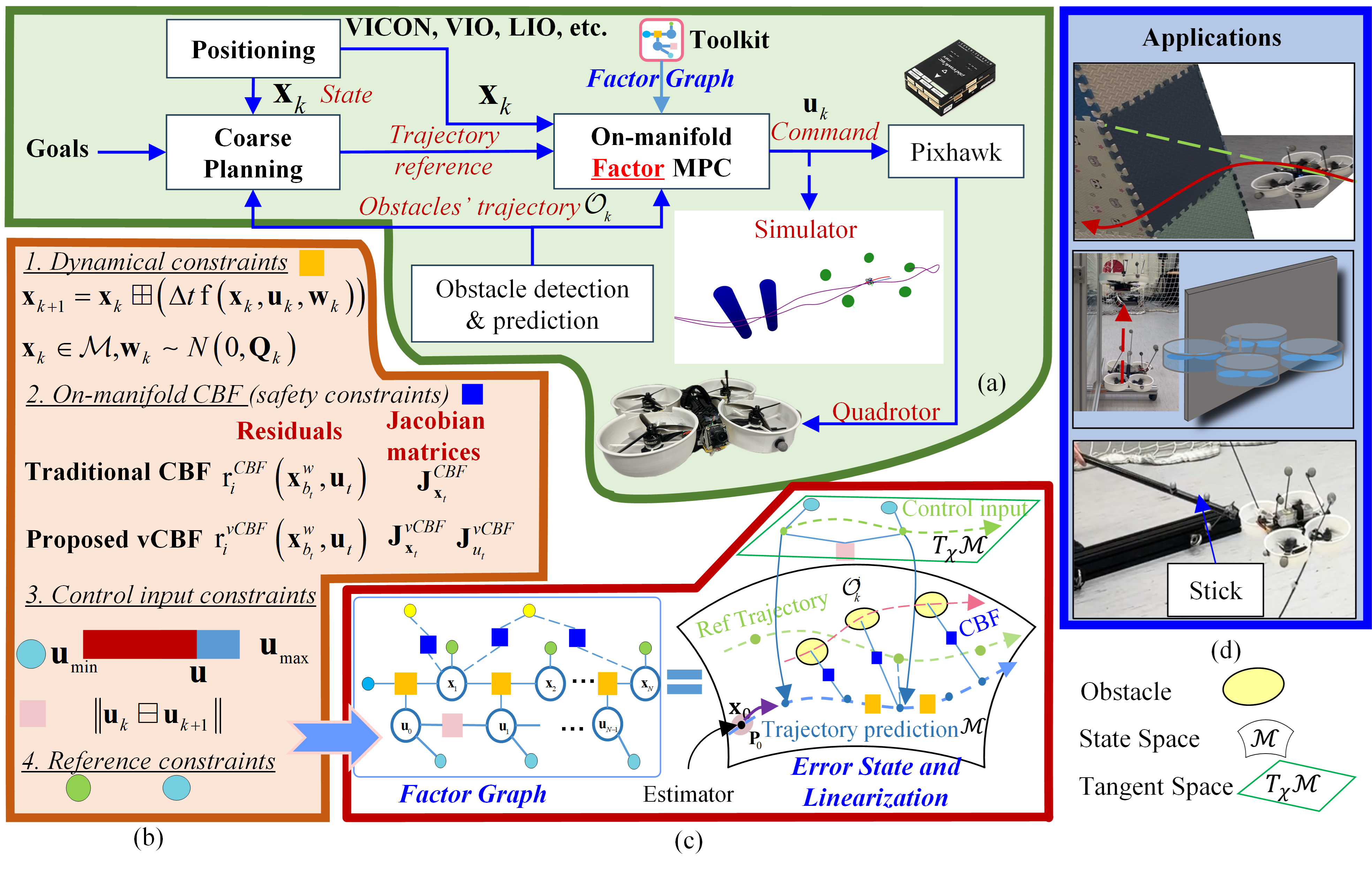}
\caption{The proposed FactorMPC framework: A unified factor graph-based toolkit for on-manifold model predictive control (MPC). (a) System architecture. (b) Key constraint formulations including dynamics, safety barriers, and state boundaries. (c) Factor graph representation of the integrated planning-control pipeline with corresponding error-state linearization details. (d) Benchmark applications demonstrating the toolkit's versatility.}
\label{fig:framework}
\end{figure*}

To address these challenges and opportunities, as shown in Fig. \ref{fig:framework}, we propose FactorMPC, an open-source toolkit for factor graph-based MPC designed for systems on manifolds. Our main contributions are:

(1) A Unified Framework: We propose a unified factor graph representation for integrated planning and control that systematically incorporates system dynamics, state constraints, and safety barriers. This framework provides explicitly derived on-manifold residuals and Jacobians, ensuring seamless integration and numerical stability for systems on nonlinear manifolds.

(2) A Novel Safety Mechanism: We introduce a novel on-manifold, velocity-extended control barrier function (CBF) framework that enhances safety in dynamic environments, providing a clear link between the CBF theory and its practical implementation within our factor graph.

(3) Open-Source Validation: We provide an open-source, highly efficient toolkit (FactorMPC) that fully leverages manifold geometry and offers modular, plug-and-play factors. The toolkit's performance is extensively validated in both simulation and physical quadrotor experiments, demonstrating superior or competitive trajectory tracking and obstacle avoidance against state-of-the-art methods.

\section{Preparations}
\subsection{Manifolds}
A manifold is a topological space that locally resembles Euclidean space, providing a framework for studying complex geometric structures in mathematics, physics, and engineering \cite{ref26}. Formally, an n-dimensional manifold $\mathcal{M}$ is a set where each point $\mathbf{x} \in \mathcal{M}$ has a neighborhood $U$ homeomorphic to an open subset of $\mathbb{R}^n$ (called the \textit{homeomorphic space}) via a bijective, continuous map $\phi: U \rightarrow \phi(U_\mathbf{x}) \subset \mathbb{R}^n$ (called \textit{homeomorphism}). These local charts enable minimal parameterization, capturing the manifold's geometry. Linear manifolds are flat, such as planes, while nonlinear manifolds, such as spheres or $SO(3)$, exhibit curvature and require multiple charts. Manifolds are crucial for modeling systems in control, robotics, and machine learning.

\subsection{System canonical representation and operators}
Consider a robotic system in discrete time with a sampling period of $\Delta t$. By applying zero-order hold discretization, where the inputs (and thus the first-order derivative of the state) remain constant over each sampling period, the system can be expressed in the following compact canonical form \cite{ref27}
\begin{equation}
\begin{aligned}
\mathbf{x}_{k+1} &= \mathbf{x}_k \boxplus (\mathrm{f} (\mathbf{x}_k, \mathbf{u}_k, \mathbf{w}_k) \Delta t) \\
\mathbf{x} &\in \mathcal{M},  \mathbf{w}_k \sim \mathcal{N}(0, \mathbf{Q}_k)
\end{aligned}
\label{eq:system_dynamics}
\end{equation}
where the state $\mathbf{x}_k$ resides on a manifold $\mathcal{M}$ of dimension $n$, with $\mathbf{u}_k \in \mathbb{R}^m$ as the system input. The vector $\mathrm{f} (\mathbf{x}_k, \mathbf{u}_k, \mathbf{w}_k) \Delta t$ represents the state perturbation induced by the input. The operation $\boxplus$ compactly denotes the ``addition'' of the state at time step $k$ and the input-induced perturbation vector, ensuring the resulting state $ \mathbf{x}_{k+1}$ remains on the manifold $\mathcal{M}$. $\operatorname{Log}$ and $\operatorname{Exp}$ represent the Logarithmic map and Exponential map of Lie theory, respectively.

Mathematically, the operator $\boxplus$ can be thought of as a map of the form: $\boxplus: \mathcal{M} \times \mathbb{R}^n \rightarrow \mathcal{M}$. Generally, $\boxplus$ depends on the specific system and its state manifold. For example, for a Lie group of dimension $n$, its canonical form could usually be cast as $\mathbf{x}_k \boxplus ( \mathrm{f} (\mathbf{x}_k, \mathbf{u}_k, \mathbf{w}_k) \Delta t )= \mathbf{x}_k \operatorname{Exp}(\mathrm{f}  (\mathbf{x}_k, \mathbf{u}_k, \mathbf{w}_k) \Delta t)$ where $\mathrm{f} (\mathbf{x}_k, \mathbf{u}_k, \mathbf{w}_k) \in \mathbb{R}^n$ lies in the same tangent space.

In the context of control systems evolving on manifolds, the $\boxplus$ and $\boxminus$ operators are instrumental in defining error states and safety constraints. For example, the error between the current state $\mathbf{x}$ and a desired state $\mathbf{x}^{r}$ can be defined as $\mathbf{e} = \mathbf{x} \boxminus \mathbf{x}^{r} \in \mathbb{R}^n $ . This error lies in the local Euclidean space, enabling the use of standard control techniques.

\subsection{Factor graphs}
Factor graphs are probabilistic graphical models composed of two node types: variable nodes for unknown states and factor nodes for the residuals between them. The cost function is to maximize the joint probability distribution defining the factors. This can be expressed by the form
\begin{equation}
p(\mathbf{X}) \propto \prod_i \exp\left(-\frac{1}{2} \| \mathrm{r}_i(\mathbf{X}) \|_{\mathbf{\Sigma}_i}^2 \right),
\label{eq:JPD}
\end{equation}
where $\mathbf{X}$ is the stacked vector containing all variable nodes, $\| \cdot \|_{\mathbf{\Sigma} }$ denotes the Mahalanobis norm, $\exp$ is the exponential function. The error functions $\mathrm{r}_i(\mathbf{X})$ are typically highly nonlinear. The maximum a posteriori estimation (MAP) objective can be reformulated as a weighted least squares problem
\begin{equation}
\mathbf{X}^{MAP} = \arg\min_{\mathbf{X}} \sum_i \| \mathrm{r}_i(\mathbf{X}) \|_{\Sigma_i}^2.
\label{eq:MAP}
\end{equation}

\section{Integrated Planning and Control}
\subsection{Model predictive control}
For a dynamic system $\dot{\mathbf{x}} = \mathrm{f} (\mathbf{x}, \mathbf{u}, \mathbf{w})$, the objective of MPC penalizes deviations between predicted states and control inputs and their respective reference values over a future time horizon $N$ \cite{ref1, ref28}. $\mathbf{x}$ is the state, including the position $\mathbf{p}$, velocity $\mathbf{v}$, and rotation $\mathbf{R}$. The optimization problem  is as follows:
\begin{equation}
\begin{aligned}
\min_{\mathbf{u}_{0:N-1}} & \mathrm{J} = \sum_{k=0}^{N-1} \left( \left\| \mathbf{x}_k - \mathbf{x}_k^r \right\|_\mathbf{Q}^2 + \left\| \mathbf{u}_k - \mathbf{u}_k^r \right\|_\mathbf{R}^2 \right) \\
&  + \left\| \mathbf{x}_N - \mathbf{x}_N^r \right\|_\mathbf{P}^2 \\
\text{s.t.} \quad &  \mathbf{x}_{k+1} = \mathbf{x}_k \boxplus \left( \Delta t \, \mathrm{f} \left( \mathbf{x}_k, \mathbf{u}_k, \mathbf{w}_k \right) \right), \\
& \mathbf{u}_{\min} \leq \mathbf{u}_k \leq \mathbf{u}_{\max},  \mathbf{u}_k \in \mathcal{U}, \\
& \mathbf{x}_k \in \mathcal{X},  k = 0, 1, \ldots, N-1, \\
& \mathbf{x}_N \in \mathcal{X}_f, \\
& \mathbf{x}_0 = \mathbf{x}_{\text{init}}
\end{aligned}
\label{eq:mpc_optimization}
\end{equation}
where $\mathbf{x}_k \in \mathcal{X}$, $\mathbf{u}_k \in \mathcal{U}$ are polyhedral, and $\mathbf{x}_N \in \mathcal{X}_f$ is a terminal polyhedral region. $\mathbf{Q} \geq 0 $, $\mathbf{R} > 0$, and $\mathbf{P} \geq 0$ are the state, input, and final state weighting matrices, respectively. $\mathbf{u}_k$ and $\mathbf{u}^{r}_k$  $(k=0,\ldots,N-1)$ is the control input and the reference input, respectively. $\mathbf{x}^{r}_k$ $(k=0,\ldots,N)$ are reference states. $\mathbf{x}_0$ is the initial state.

\subsection{Control barrier functions}
CBF is a function $\mathrm{h}( \mathbf{x} )$ defined on the state space of a system, where the safe set $\mathcal{C}$ in the state space where the system should remain, is represented by
\begin{equation}
\mathcal{C} = \{ \mathbf{x} \in \mathbb{R}^n : \mathrm{h} ( \mathbf{x} ) \geq 0 \},
\label{eq:cbf_set}
\end{equation}
where $\mathrm{h}(\mathbf{x})$ is a continuously differentiable function. Lie derivative condition: For the system $\dot{\mathbf{x}} = \mathrm{f}(\mathbf{x}, \mathbf{u})$, a function $\mathrm{h}(\mathbf{x})$ is a CBF if there exists a control law such that the time derivative of $\mathrm{h}(\mathbf{x})$ satisfies the constraint.
\begin{equation}
\dot{\mathrm{h} }( \mathbf{x}, \mathbf{u} ) \geq -\mathrm{f}_\alpha( \mathrm{h} ( \mathbf{x} )),
\label{eq:cbf_cons}
\end{equation}
where $\mathrm{f}_\alpha(\cdot)$ is a $\mathcal{K}$ function that shapes the behavior of the system near the boundary of the safe set, and $\dot{\mathrm{h} }(\mathbf{x}, \mathbf{u}) = L_f \mathrm{h}(\mathbf{x}) + L_g \mathrm{h}(\mathbf{x}) \mathbf{u}$. $L_f \mathrm{h}(\mathbf{x})$ is Lie derivatives representing the rate of change of $\mathrm{h}( \mathbf{x})$ along the system dynamics $\mathrm{f}(\mathbf{x})$. CBFs are often integrated with optimization-based control, such as MPC. Integrating (\ref{eq:mpc_optimization}) and (\ref{eq:cbf_cons}), the safety-critical MPC based on CBF constraints is to solve the problem
\begin{equation}
\begin{aligned}
\min_{\mathbf{u}_{0:N-1}} & \mathrm{J}  \\
\text{s.t.} \quad & (\ref{eq:mpc_optimization})'s  \quad constraints \\
& \dot{\mathrm{h} }(\mathbf{x}_k, \mathbf{u}_k) \geq -\mathrm{f}_\alpha(\mathrm{h}(\mathbf{x}_k) )
\end{aligned}
\label{eq:mpc_cbf}
\end{equation}

Note that, from this point, we will make the common assumption that $\mathrm{f}_\alpha (\mathrm{h}(\cdot) ) := \alpha \mathrm{h}(\cdot), \alpha \in \mathbb{R} > 0$.
\subsection{Factor graph-based unified problem}
\begin{figure}[thpb]
\centering
\includegraphics[]{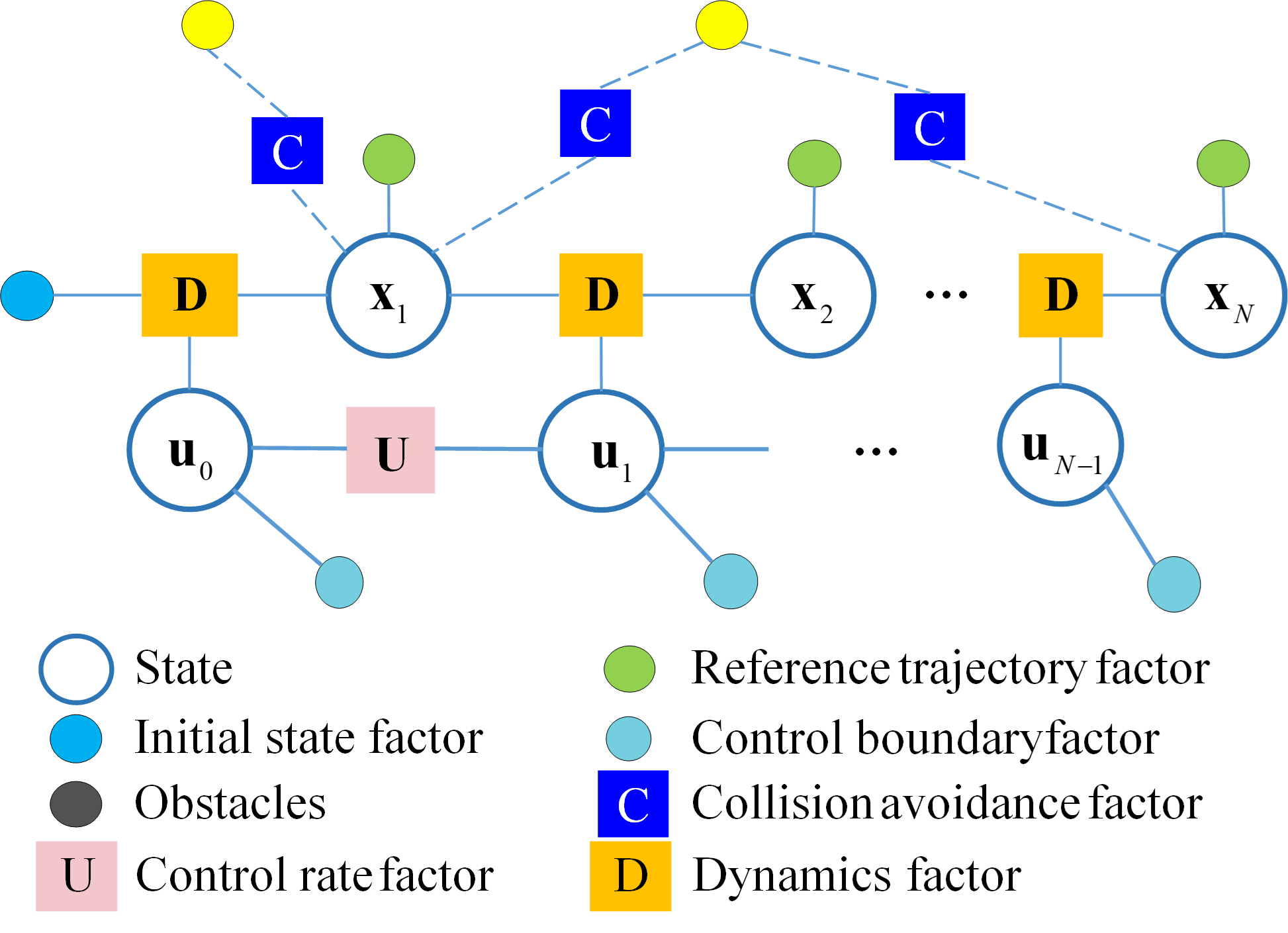}
\caption{The factor graph of MPC.}
\label{fig:factor_graph}
\end{figure}

To represent the MPC using factor graphs, Eq. (\ref{eq:mpc_cbf}) is reformulated as a factor graph optimization problem.
\begin{equation}
\begin{aligned}
\max_{\mathbf{u}_{0:N-1}} & \left( 
\begin{aligned}
& \sum_{k=0}^{N-2} \left\| \mathrm{r}^U \left( \mathbf{u}_k, \mathbf{u}_{k+1} \right) \right\|_\mathbf{R}^2 
+ \sum_{k=0}^{N-1} \left\| \mathbf{r}^B \left( \mathbf{\mathbf{u}_k} \right) \right\|_{\mathbf{Q}_B}^2 \\
& + \sum_{k=1}^{N-1} \left\| \mathrm{r}^D \left( \mathbf{\cdot} \right) \right\|_\mathbf{P}^2 
+ \sum_{k=0}^{N-1} \sum_{i=0}^{N-1} \left\| \mathrm{r}_i^{CBF} \left(\cdot \right) \right\|^2 \\
& + \sum_{k=1}^{N} \left\| \mathrm{r}^{ref} \left( \mathbf{x}_k \right) \right\|_\mathbf{Q}^2 
+ \left\| \mathbf{x}_0 \boxminus \hat{\mathbf{x}}_{\text{init}} \right\|_{\mathbf{P}_0}^2 
\end{aligned}
\right)
\end{aligned}
\label{eq:FG_problem}
\end{equation}
where $\mathbf{R}$, $\mathbf{Q}_B$, $\mathbf{P}$, and $\mathbf{Q}$ represent different weighting matrices. $\mathbf{P}_0$ is the covariance matrix of $\mathbf{x}_0$. The factor graph in Fig. \ref{fig:factor_graph} represents the structure of the optimization problem (\ref{eq:FG_problem}). The details of these factors are explained below.
\subsubsection{Universal dynamics factor}

A typical dynamic model of an autonomous vehicle is as follows:
\begin{equation}
\begin{aligned}
\dot{\mathbf{p}}^w_b &= \mathbf{v}^w_b, \\
\dot{\mathbf{R}}^w_b  &= \mathbf{R}^w_b \omega_b^\times, \\
\dot{\mathbf{v}}^w_b &=\mathbf{R}^w_b \mathbf{a}_b - g \mathbf{e}_3,
\end{aligned}
\label{eq:dynamic_model}
\end{equation}
where $\mathbf{a}_b$ is the acceleration. $\omega_b$ is the angular velocity. $g$ is the gravitational constant. $\mathbf{e}_3 = [0,0,1]^T$.

The dynamics factor residual \cite{ref25} is given by
\begin{equation}
{\small
\begin{aligned}
& \mathrm{r}^D \left(\mathbf{x}_{b_k}^{w}, \mathbf{u}_{k}, \mathbf{x}_{b_{k+1}}^{w}\right) \\
& =
\begin{bmatrix}
\mathbf{R}_{w}^{b_k}\left(\mathbf{p}_{b_{k+1}}^{w}-\mathbf{v}_{b_k}^{w} \Delta t + 0.5\mathbf{e}_3 g \Delta t^2-\mathbf{p}_{b_k}^{w}\right) - 0.5\mathbf{a}_{b_k}\Delta t^2 \\
\operatorname{Log}\left(\mathbf{R}_{w}^{b_k}\mathbf{R}_{b_{k+1}}^{w}\right)-\boldsymbol{\omega}_{b_k}\Delta t \\
\mathbf{R}_{w}^{b_k}\left(\mathbf{v}_{b_{k+1}}^{w}-\mathbf{v}_{b_k}^{w}+\mathbf{e}_3 g \Delta t\right)-\mathbf{a}_{b_k}\Delta t
\end{bmatrix}
\label{eq:dynamic_model_res}
\end{aligned} }
\end{equation}
where $\Delta t$ is the period between two contiguous states. 

Furthermore, the covariance matrix $\mathbf{P}$ of $\mathrm{r}^D$ is defined by matrix [25]
\begin{equation}
\mathbf{P} = \Delta t^2
\begin{bmatrix}
0.25\Delta t^2\sigma_a^2 \mathbf{I}_3 & \mathbf{0} & 0.5\Delta t\sigma_a^2 \mathbf{I}_3 \\
\mathbf{0} & \sigma_\omega^2 \mathbf{I}_3 & \mathbf{0} \\
0.5\Delta t\sigma_a^2 \mathbf{I}_3 & \mathbf{0} & \sigma_a^2 \mathbf{I}_3
\end{bmatrix}
\label{eq:covariance_matrix}
\end{equation}
where $\sigma_a$ and $\sigma_\omega$ are the Gaussian noise standard deviations for the linear acceleration and angular velocity.

\subsubsection{Control boundary and control rate factor}

The control boundary factor ensures that the control input meets the actuators' limits. We define a bounding function  $\mathrm{r}^B(\mathbf{u}_k)$ for the control input inequality as follows: 
\begin{equation}
\mathrm{r}^B(\mathbf{u}_k) = \max{(\mathbf{u}_k - \mathbf{u}_{max}, \mathbf{0} ) } + \max{(\mathbf{u}_{min} - \mathbf{u}_k, \mathbf{0} ) }
\label{eq:cbcr_res}
\end{equation}
where $\mathbf{u}_{min}$ is the control input lower bound, $\mathbf{u}_{max}$ is the control input upper bound, and $\max{()}$ is the element-wise maximum operator.

To smooth the trajectory, the control rate factor is used to limit the rate of control inputs, which is as follows:
\begin{equation}
\mathrm{r}^U(\mathbf{u}_k, \mathbf{u}_{k-1}) = \mathbf{u}_k - \mathbf{u}_{k-1}.
\label{eq:crf_res}
\end{equation}

\subsubsection{Reference trajectory factor} 
The reference trajectory factor's residual \cite{ref25} is as follows:
\begin{equation}
\mathrm{r}^{ref}( \mathbf{x}_k, \mathbf{x}^{r}_k) = \mathbf{x}_k \boxminus \mathbf{x}^r_k.
\label{eq:rtf_res}
\end{equation}

\subsubsection{Distance-based CBF collision avoidance factor}

\begin{figure}[thpb]
\centering
\begin{subfigure}{0.30\textwidth}
    \centering
    \includegraphics[width=\linewidth]{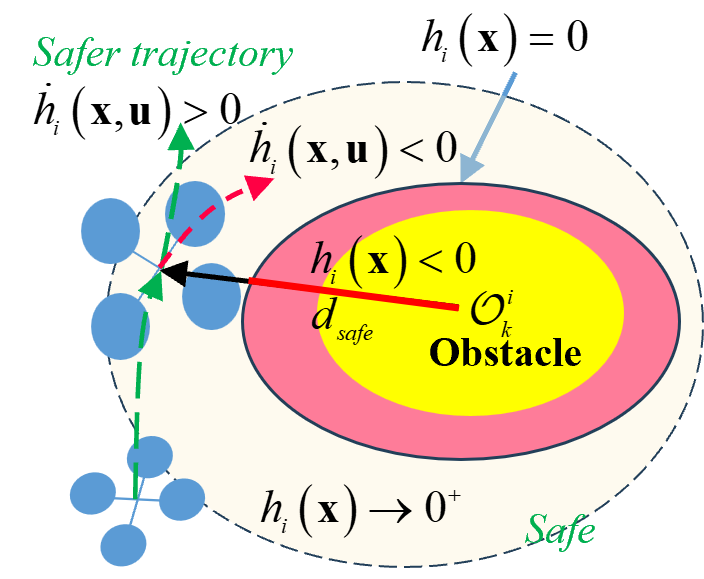}
    \caption{}
    \label{fig:3a}
\end{subfigure}
\hfill
\begin{subfigure}{0.30\textwidth}
    \centering
    \includegraphics[width=\linewidth]{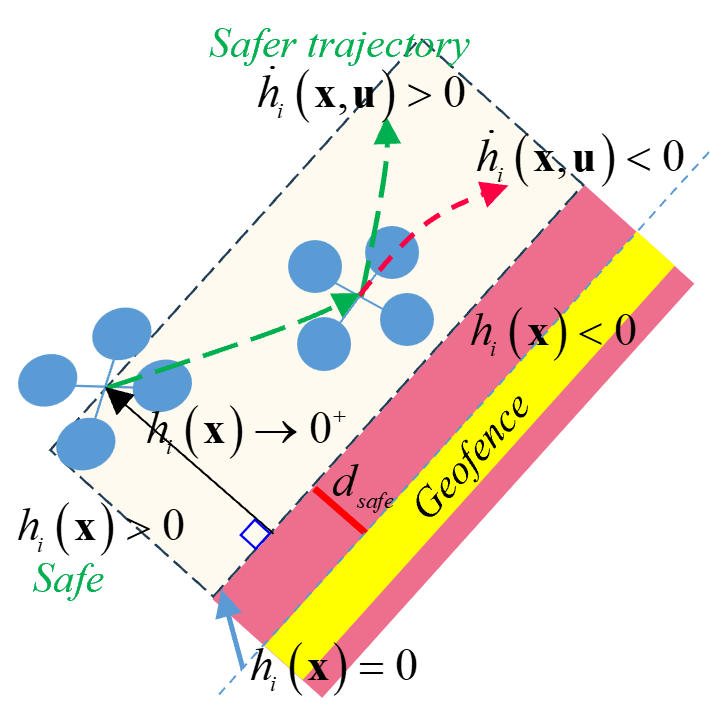}
    \caption{}
    \label{fig:3b}
\end{subfigure}
\caption{The illustration of CBFs. (a) Sphere obstacle. (b) Geofence.}
\label{fig:cbf_figs}
\end{figure}

For a spherical obstacle or geofence, a distance-based CBF defines the safe set. As shown in Fig. \ref{fig:cbf_figs}, when the vehicle is near the obstacle or geofence boundary ($\mathrm{h}( \mathbf{x} ) = 0$), the CBF constraint $\dot{\mathrm{h} }(\mathbf{x},  \mathbf{u} ) \geq -\alpha(\mathrm{h}(\mathbf{x}))$ is enforced to modify the control input, ensuring the vehicle's state remains in the safe set for all time. The CBF guarantees forward invariance of the safe set.
Distance-based safe set that measures the Euclidean distance between the robot's position and the obstacles (or critical points on its body)
\begin{equation}
\begin{aligned}
S = \left\{ \mathbf{x} \in \mathbb{R}^n \middle| \mathrm{h}_i^{CBF} \left( \mathbf{x}_{b}^{w}, t \right) = \frac{1}{d_{safe}} - \frac{1}{d_{o}^{i,t}} \geq 0, \right. \\
\left. d_{o}^{i,t} = \left\| \mathbf{p}_{o}^{i,t} \right\| = \left\| \mathbf{p}_{b}^{w} - \mathbf{p}_{o_i^t}^{w} \right\|, 0 \leq i \leq S-1 \right\}
\end{aligned}
\label{eq:distance_set}
\end{equation}
where  $\mathbf{p}_{o_i^t}^{w}$  represents the position of the \( i \)-th obstacle of \( S \)-obstacles at the timestamp \( t \) and \( d_{safe} \) represents the safe distance between the MAVs and obstacles. \( d_{safe} \) is set as the radius of the \( i \)-th obstacle plus the safety margin (\( d_{safe} \) = \( d_{radius}^i \) + \( d_{margin} \)). The dynamics model (\ref{eq:dynamic_model}) can be rewritten as an affine form
\begin{equation}
\begin{aligned}
& \dot{\mathbf{x} } = \mathrm{f}(\mathbf{x} ) + \mathrm{g} (\mathbf{x} ) \mathbf{u}, \\
& \mathrm{f} (\mathbf{x} ) = 
\begin{bmatrix}
\mathbf{\mathbf{v} } \\
\mathbf{0} \\
-\mathbf{e}_3 g \\
\end{bmatrix},  \mathrm{g} (\mathbf{x} ) = 
\begin{bmatrix}
\mathbf{0} & \mathbf{0} \\
\mathrm{H} (\boldsymbol{\theta})^{-1} & \mathbf{0} \\
\mathbf{0} & \mathbf{R} \\
\end{bmatrix},  \mathbf{u} = 
\begin{bmatrix}
\boldsymbol{\omega} \\
\mathbf{a} \\
\end{bmatrix}
\label{eq:affine_form}
\end{aligned}
\end{equation}
where function \( \mathrm{H} (\boldsymbol{\theta}) \) is given by a formula
\begin{equation}
\mathrm{H} (\boldsymbol{\theta}) = \sum_{k=0}^{\infty} \frac{(-1)^k}{(k+1)!} \left( \left[ \boldsymbol{\theta} \right]_{\times}^k \right),  \boldsymbol{\theta} = \text{Log}(\mathbf{R})
\label{eq:17}
\end{equation}

According to \eqref{eq:distance_set}, \eqref{eq:affine_form}, and  \eqref{eq:cbf_cons}, the safety constraints based on distance-based CBF are as follows:
\begin{equation}
\begin{aligned}
& \alpha \mathrm{h}^{CBF}_i(\mathbf{x}^w_b,t) + \dot {\mathrm{h}}^{CBF}_i(\mathbf{x}^w_b,t) \\ & = \alpha \left( \frac{1}{d_{\text{safe}}} - \frac{1}{d_o^{i,t}} \right) + \frac{(\mathbf{p}_o^{i,t})^T}{(d_o^{i,t})^3} \mathbf{v}_b^w \geq 0   
\end{aligned}
\end{equation}

The distance-based CBF's residuals $\mathrm{r}_i^{\mathrm{CBF}}(\mathbf{x}_{b_k}^w)$ are defined by,
\begin{equation}
\begin{aligned}
& \mathrm{r}_i^{\mathrm{CBF}}(\mathbf{x}_{b_k}^w) \\
& = \max \left( -\alpha \left( \frac{1}{d_{\text{safe}}} - \frac{1}{d_o^{i,k}} \right) - \frac{(\mathbf{p}_o^{i,k})^T}{(d_o^{i,k})^3} \mathbf{v}_{b_k}^w, 0  \right) 
\end{aligned}
\end{equation}
where symbol abuses and \( k \Delta t \) is directly replaced by $k$.
\subsubsection*{5) Velocity-extended CBF collision avoidance factor}
Furthermore, relative velocity in the direction of the inter-centroid vector is considered as an additional constraint. Therefore, for collision avoidance, the velocity-extended CBF ($\mathrm{vCBF}$)
\begin{equation}
\mathrm{h}_i^{\mathrm{vCBF}}(\mathbf{x}_b^w, t) = \frac{1}{d_{\text{safe}}} - \frac{1}{d_o^{i,t}} + \gamma \mathbf{n}_i^T(t) (\mathbf{v}_b^w - \mathbf{v}_{o_t^i}^w)
\label{eq:vCBF}
\end{equation}
where $\mathbf{v}_{o_t^i}^w$ is the $i$-th obstacle velocity at the timestamp $t$, $\gamma$ is a constant coefficient, and $\mathbf{n}_i(t) = (\mathbf{p}_b^w - \mathbf{p}_{o_t^i}^w)/\|\mathbf{p}_b^w - \mathbf{p}_{o_t^i}^w\|$.

Then, $\dot{\mathrm{h}}_i^{\mathrm{vCBF}}(\mathbf{x}_b^w, t)$ is as follows:
\begin{equation}
\begin{aligned}
& \dot{\mathrm{h}}_i^{\mathrm{vCBF}}(\mathbf{x}_b^w, t, \mathbf{u}_t)   = \frac{\partial \mathrm{h}_i^{\mathrm{vCBF}}(\mathbf{x}_b^w, t)}{\partial \mathbf{x}_b^w} \mathbf{x}_b^w \\ 
& = \left[ \frac{\partial \mathrm{h}_i^\mathrm{vCBF}}{\partial \mathbf{p}_b^w}, \mathbf{0}^T_3, \frac{\partial \mathrm{h}_i^\mathrm{vCBF}}{\partial \mathbf{v}_b^w} \right] \left\{ \mathrm{f}(\mathbf{x}_b^w) + \mathrm{g}(\mathbf{x}_b^w) \mathbf{u}_t \right\} 
\end{aligned}
\label{eq:dotvCBF}
\end{equation}

According to (\ref{eq:vCBF}) and (\ref{eq:dotvCBF}), $\mathbf{r}_i^{\mathrm{vCBF}}(\mathbf{x}_{b_k}^w, \mathbf{u}_k)$ is as follows:
\begin{equation}
\begin{aligned}
& \mathbf{r}_i^{\mathrm{vCBF}}(\mathbf{x}_{b_k}^w, \mathbf{u}_k)
 \\ &= \max \left\{  - \frac{\partial \mathrm{h}_i^{\mathrm{vCBF}}}{\partial \mathbf{p}_{b_k}^w} \mathbf{v}_{b_k}^w - \frac{\partial \mathrm{h}_i^{\mathrm{vCBF}}}{\partial \mathbf{v}_{b_k}^w} (-\mathbf{e}_3 g + \mathbf{R}_{b_k}^w \mathbf{a}_{b_k}) \right. \\
& \left. - \alpha \mathrm{h}_i^{\mathrm{CBF}}(\mathbf{x}_{b_k}^w), 0 \right\}
\end{aligned}
\end{equation}

The Jacobian matrix $\mathbf{J}_{u}^i$ of $\mathbf{r}_i^{\mathrm{vCBF}}(\mathbf{x}_{b}^w, \mathbf{u})$ with $\mathbf{u}$ when $\dot{\mathrm{h}}^{\mathrm{vCBF}}_i + \alpha \mathrm{h}^{\mathrm{vCBF}}_i < 0$  is as follows:
\begin{equation}
\begin{aligned}
\mathbf{J}_{u}^i & = \frac{\partial \mathrm{r}_i^{\mathrm{vCBF}}(\mathbf{x}_{b}^w, \mathbf{u})}{\partial \mathbf{u}} \\
& = \frac{\partial \left[ \frac{\partial \mathrm{h}_i^{\mathrm{vCBF}}}{\partial \mathbf{v}_b^w} (\mathbf{e}_3 g + \mathbf{R}_b^w \mathbf{a}_b) \right]}{\partial \mathbf{u}} 
\end{aligned}
\end{equation}

\subsection{Quadrotor case}
The dynamic model of the quadrotor \cite{ref29} is as follows:
\begin{equation}
\begin{aligned}
\dot{\mathbf{p}}_b^w &= \mathbf{v}_b^w \\
\dot{\mathbf{R}}_b^w &= \mathbf{R}_b^w \boldsymbol{\omega}_b^\times \\
\dot{\mathbf{v}}_b^w & = -\mathbf{e}_3 g + \frac{\mathbf{R}_b^w \mathbf{e}_3 T_b}{m_b} + \frac{\mathbf{R}_b^w \mathbf{D} \mathbf{R}_b^w \mathbf{v}_b^w}{m_b}
\end{aligned}
\label{eq:quadrotor_dynamics}
\end{equation}
where $T_b$ is the thrust command, $m_b$ is the mass, and $\mathbf{D} \in \mathbb{R}^{3 \times 3}$ represents the drag matrix. For the quadrotor model, $\mathbf{u} = [\boldsymbol{\omega}_b, T_b]^T$, the Jacobian $\mathbf{J}_{u_k}^i $ becomes,
\begin{equation}
\begin{aligned}
\mathbf{J}_{u_k}^i  & = \frac{\partial \gamma \alpha \mathbf{n}_i(k) \left( \frac{\mathbf{R}_{b_k}^w \mathbf{e}_3 T_{b_k}}{m_b} \right)}{\partial \mathbf{u}_k} \\
& = \begin{bmatrix} \mathbf{0}_3^T & \gamma \alpha \mathbf{n}_i(k) \mathbf{R}_{b_k}^w \mathbf{e}_3 / m_b \end{bmatrix}^T \\
& = \begin{bmatrix} \mathbf{0}_3^T & \gamma \alpha \cos(\theta_k^i) / m_b \end{bmatrix}^T
\label{eq:jacobian_u}
\end{aligned}
\end{equation}
where $\theta_k^i$ is the angle between $\mathbf{n}_i(k)$ and $\mathbf{R}_{b_k}^w \mathbf{e}_3$. $\mathbf{R}_b^w \mathbf{e}_3$ is a vector representing the $z$-axis of the body frame expressed in the world frame. The Levenberg-Marquardt algorithm update is as follows:
\begin{equation}
\begin{aligned}
\Delta \mathbf{u}^i_k & = -\left( \left(\mathbf{J}_{u_k}^i \right)^T \mathbf{J}_{u_k}^i + \lambda \mathbf{I} \right)^{-1} \left(\mathbf{J}_{u_k}^i \right)^T \mathbf{r}_i^{\text{vCBF}}(\mathbf{x}_{b_k}^w, \mathbf{u_k}) \\ 
& \left( \left(\mathbf{J}_{u_k}^i \right)^T \mathbf{J}_{u_k}^i + \lambda \mathbf{I} \right)^{-1} \left(\mathbf{J}_{u_k}^i \right)^T \\
& = \left( \gamma^2 \alpha^2 \cos^2(\theta^i_k) / m_b^2 + \lambda \right)^{-1} \begin{bmatrix} \mathbf{0}_3 \\ \gamma \alpha \cos(\theta^i_k) / m_b \end{bmatrix}
\label{eq:lm_update}
\end{aligned}
\end{equation}
where the non-negative damping factor $\lambda$ is adjusted at each iteration. When $\theta^i = \pi / 2$, the $\lambda$ is used to mitigate the ill-conditioned matrix.
\subsection{Toolkit development}

To widen the adaptability of the FactorMPC\footnote{All Jacobian matrices are introduced with details in supplementary materials at: \url{https://github.com/RoboticsPolyu/FactorMPC}.}
, the proposed toolkit adopts a universal dynamics model that outputs thrust and angular velocity control variables, which are then tracked by low-level controllers. This makes it easier to port the proposed MPC to different platforms without the need for repeated development. We open-source all factors and provide examples for quadrotors. The toolkit is developed based on Gtsam \cite{ref23}. All factors can be inserted into the iSAM \cite{ref23} solver. 

\section{Simulation and Experiments}

\subsection{Simulation scheme and results}

This study conducts a series of simulations to evaluate and compare the efficacy of advanced motion planning and safety-critical control frameworks for a quadrotor. First, we benchmark the performance of our proposed method against a baseline gradient-based planner, EGO-Planner \cite{ref30}. Second, to guarantee safety amidst obstacles, we compare two CBF formulations: a standard distance-based CBF and a more dynamic velocity-extended CBF (vCBF). Finally, the culmination of these tests is a comprehensive multi-obstacle path planning simulation, where the optimal planner with the robust CBF-based factors is rigorously validated in cluttered 3D spaces.

\subsubsection{Static obstacle avoidance}

\begin{figure}[htbp]
    \centering
    \includegraphics[]{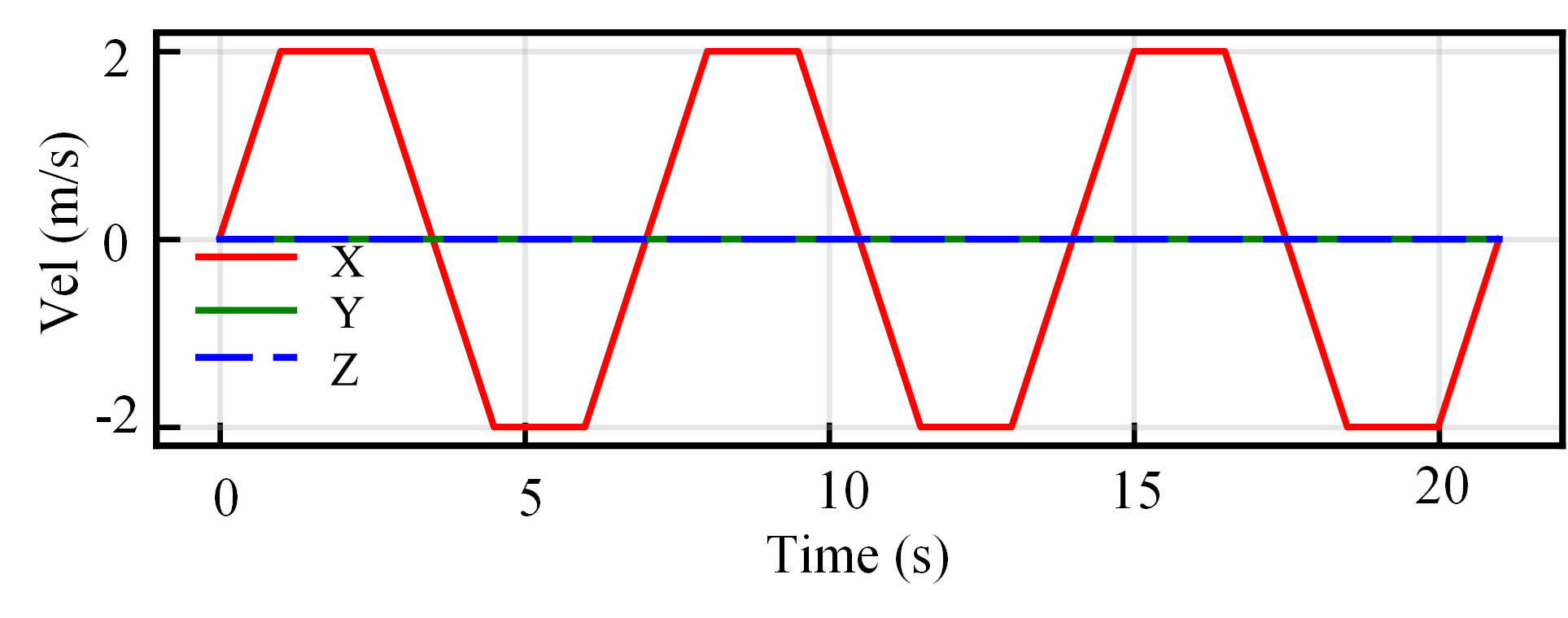}
    \caption{The quadrotor velocity reference curve.}
    \label{fig:velocity_profile}
\end{figure}

\begin{figure}[htbp]
    \centering
    \begin{subfigure}{0.9\columnwidth}
        \centering
        \includegraphics[width=\textwidth]{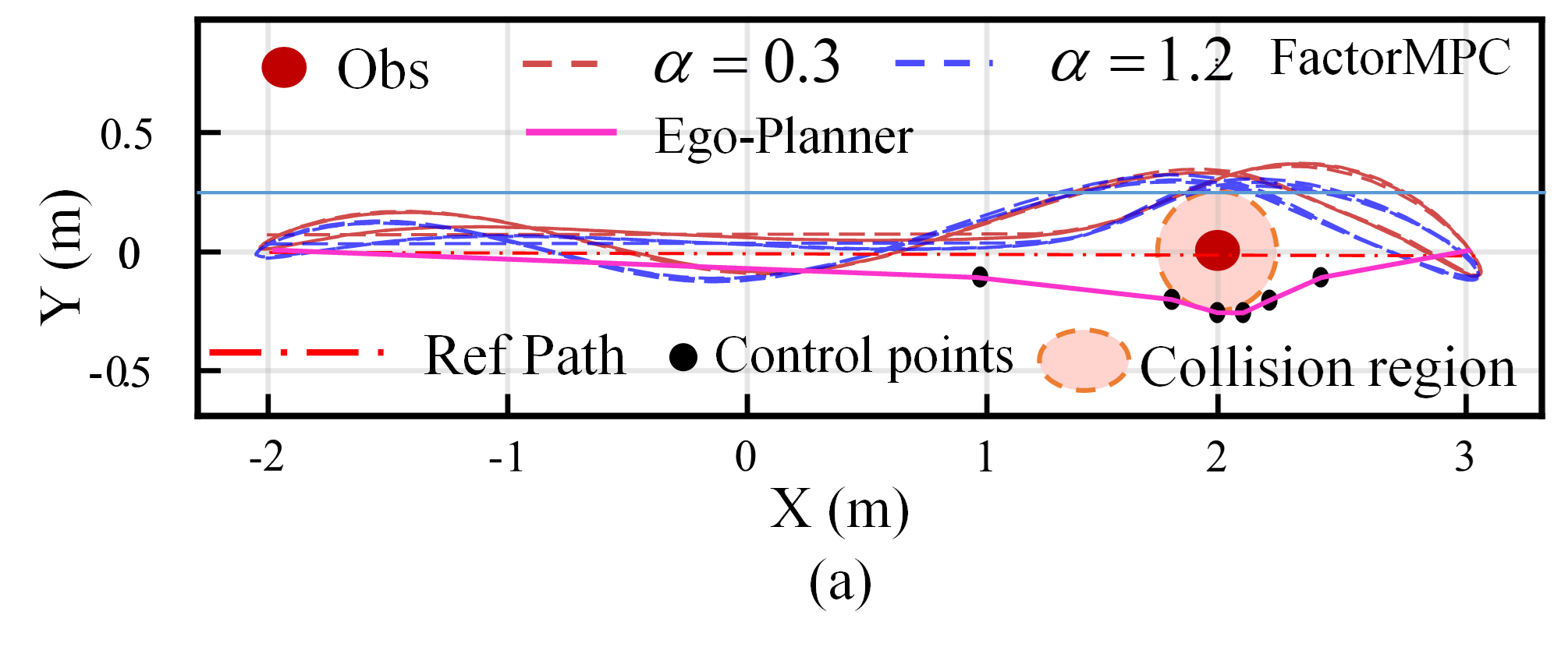}
    \end{subfigure}
    \hfill
    \begin{subfigure}{0.9\columnwidth}
        \centering
        \includegraphics[width=\textwidth]{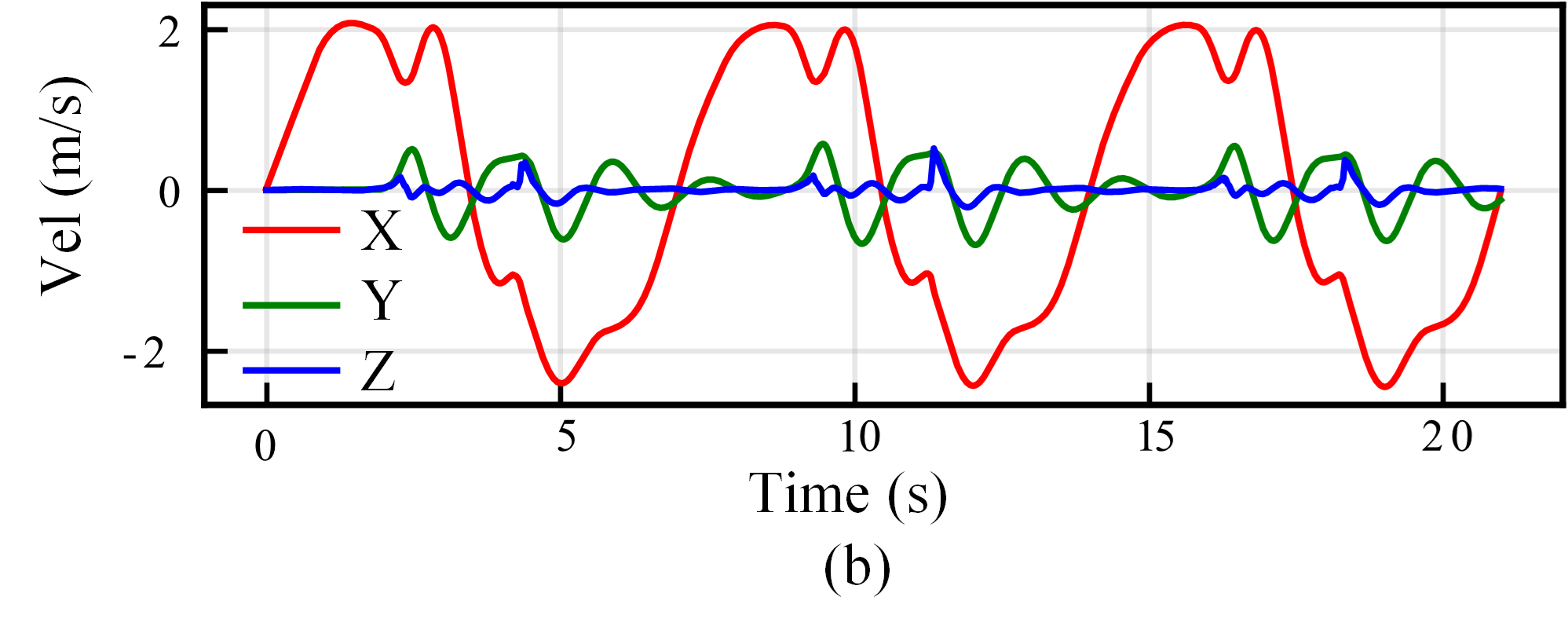}
    \end{subfigure}
    \\
    \begin{subfigure}{0.95\columnwidth}
        \centering
        \includegraphics[width=\textwidth]{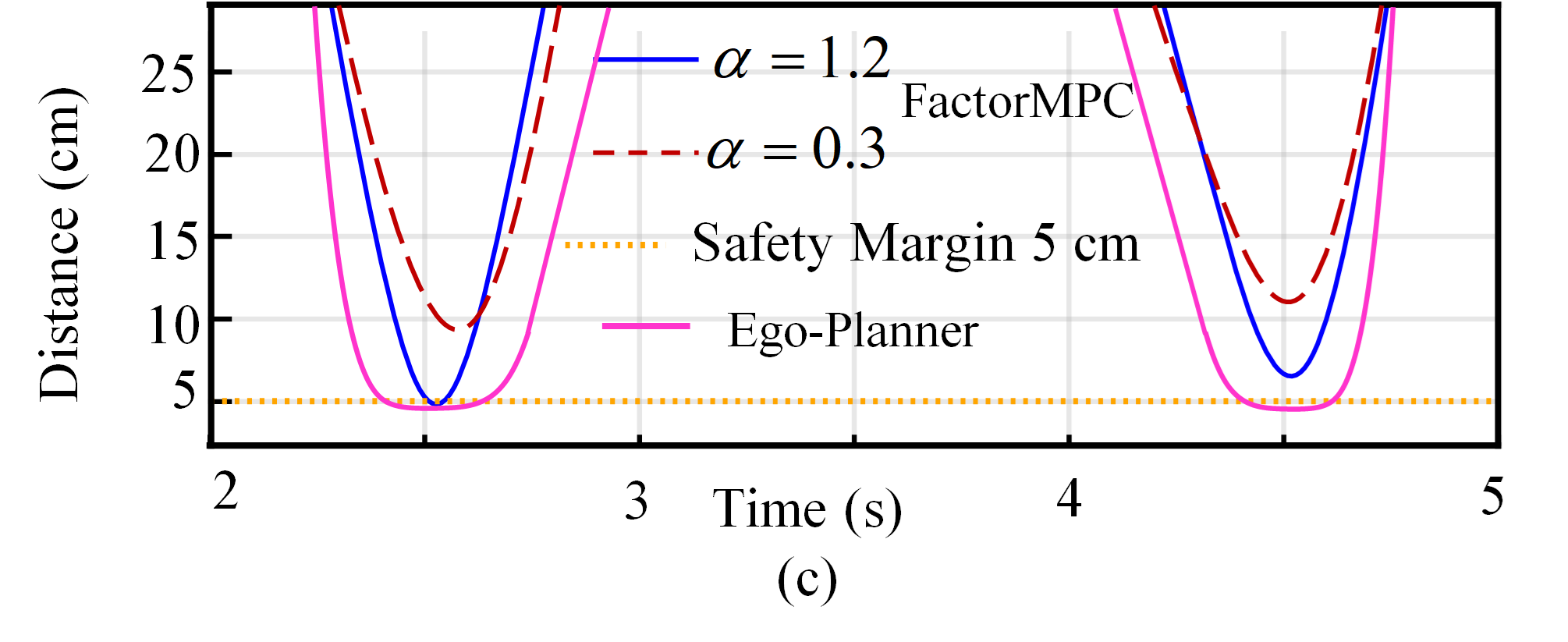}
    \end{subfigure}
    \caption{(a) The quadrotor’s path using distance-based CBF under different conditions: blue dashed line depicts $\alpha  = 1.2 $ and purple solid line depicts $\alpha = 0.3$. (b) The quadrotor’s velocity when $\alpha = 0.3 $. (c) The boundary distance between the obstacle and quadrotor.}
    \label{fig:cbf_variation}
\end{figure}

The scenario involves a static cylindrical obstacle with a straight-line reference trajectory. Fig.~\ref{fig:velocity_profile} illustrates the resulting velocity profiles, while Fig.~\ref{fig:cbf_variation} demonstrates the effect of varying the parameter $\alpha$ in the distance-based CBF framework. As $\alpha$ decreases, the system exhibits greater safety margin, maintaining a larger minimum distance from the obstacle. This trade-off between safety margin and tracking performance highlights the critical role of $\alpha$ in balancing conservatism and agility in obstacle avoidance. The comparative results in Fig.~\ref{fig:cbf_variation}(a) and Fig.~\ref{fig:cbf_variation}(b) indicate better safety performance of the proposed method over EGO-Planner when using equivalent safety margins. The susceptibility of EGO-Planner to enter unsafe regions is attributed to its dependence on optimizing B-spline control points. To mitigate this, a hierarchical framework is suggested: EGO-Planner acts as an initial high-level planner to generate a coarse trajectory, which is then processed by the proposed method—functioning as a certifying safety filter—to ensure strict adherence to safety constraints.

\subsubsection{Distance-based CBF versus velocity-extended CBF}

\begin{figure}[htbp]
    \centering
    \begin{subfigure}{0.9\columnwidth}
        \centering
        \includegraphics[width=\textwidth]{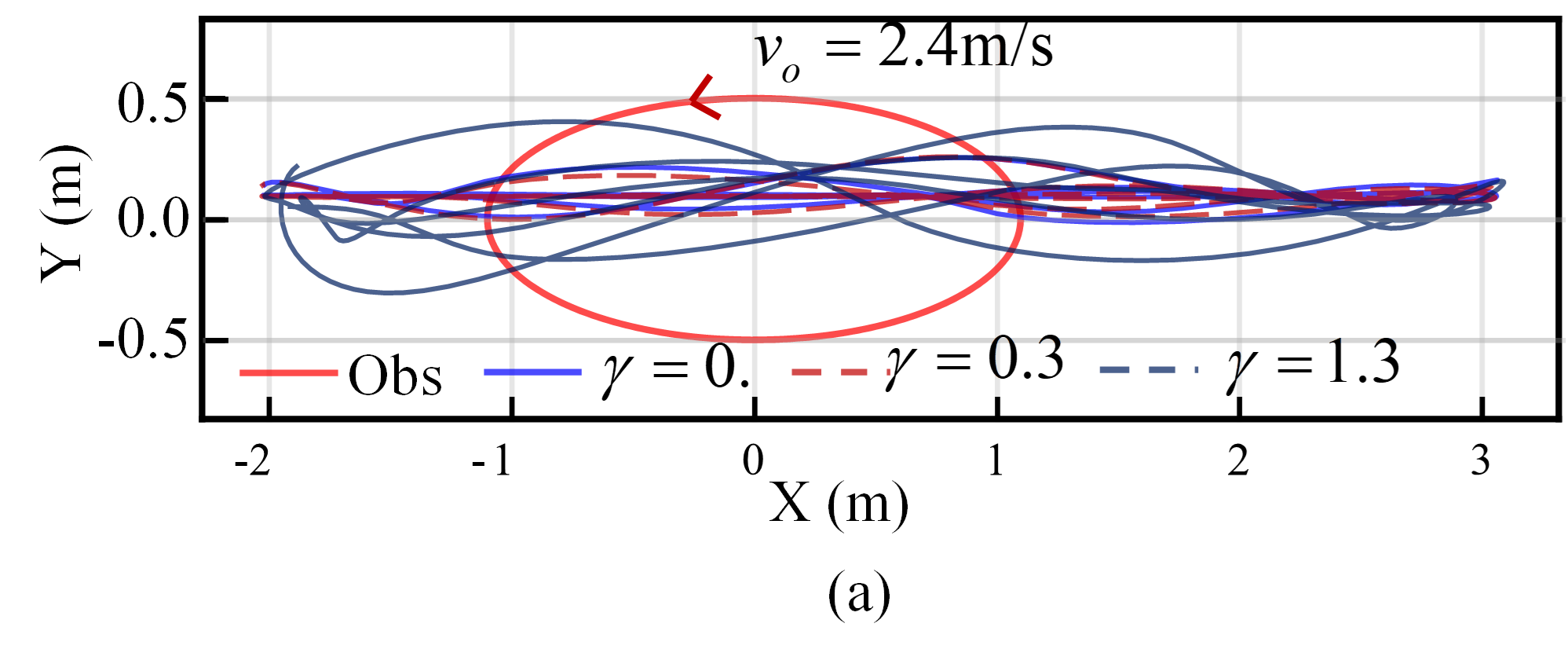}
    \end{subfigure}
    \hfill
    \begin{subfigure}{0.9\columnwidth}
        \centering
        \includegraphics[width=\textwidth]{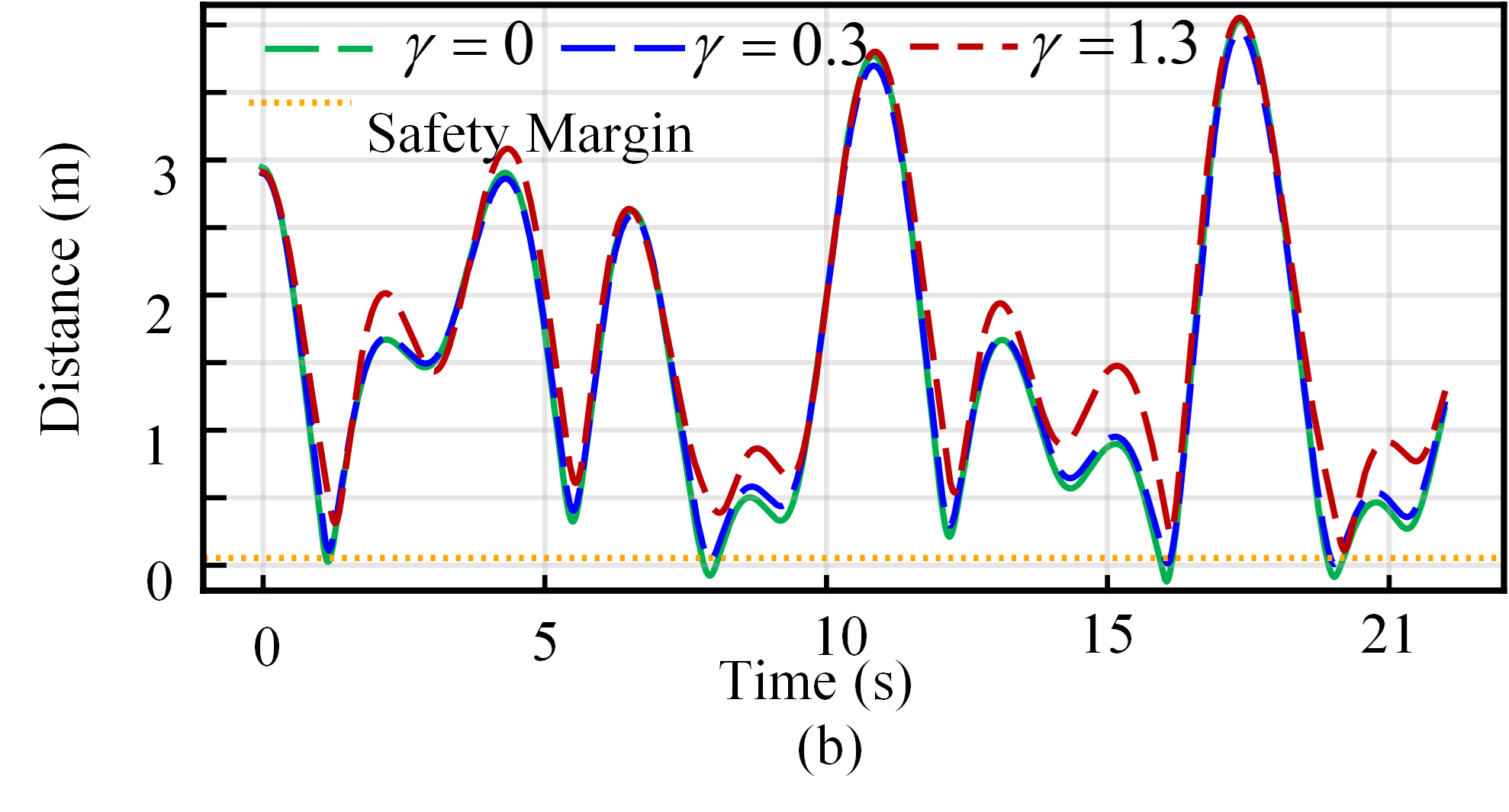}
    \end{subfigure}
    \\
    \begin{subfigure}{0.9\columnwidth}
        \centering
        \includegraphics[width=\textwidth]{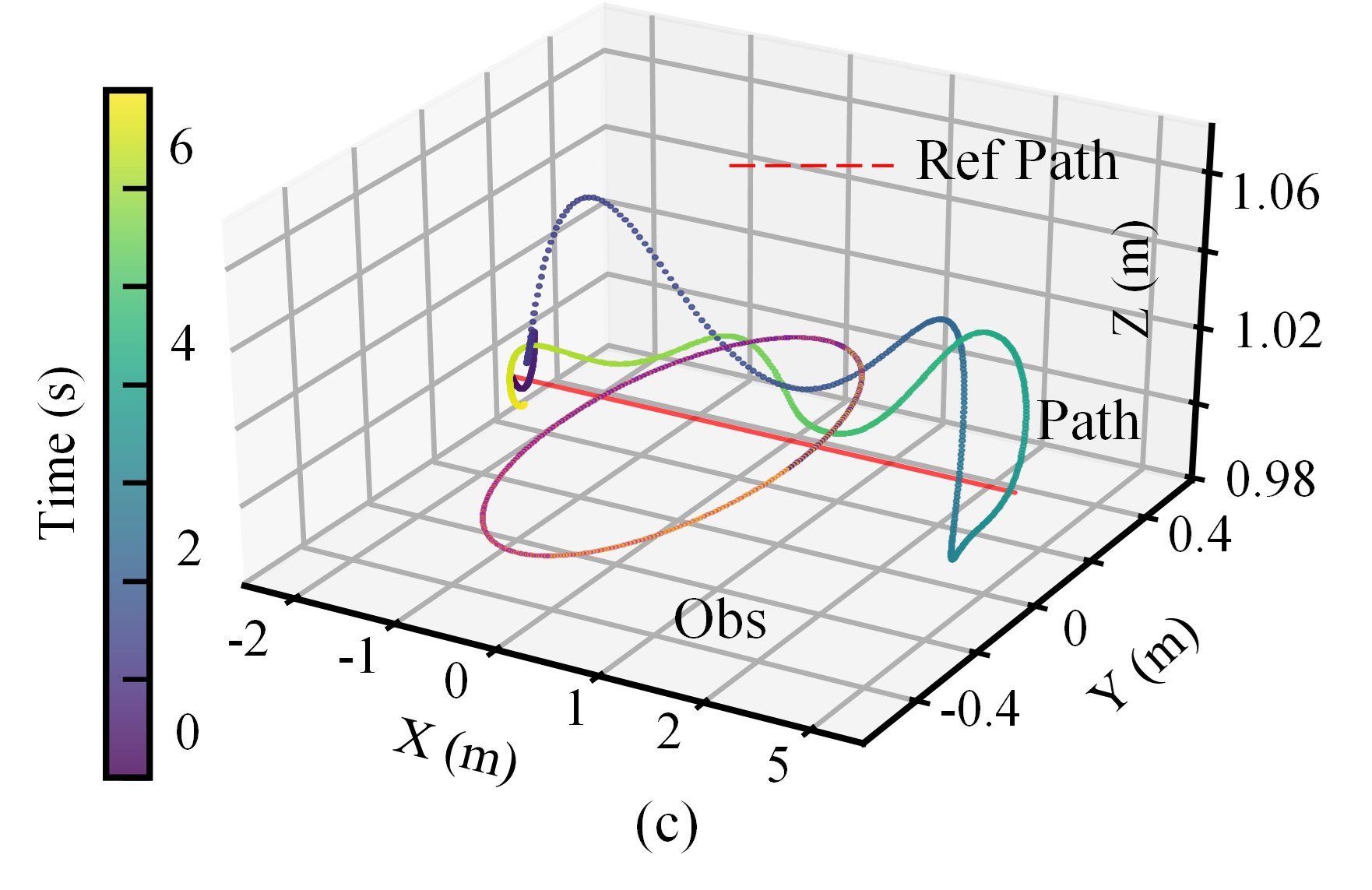}
    \end{subfigure}
    \caption{(a) Obstacles with uniform elliptical motion, obstacle avoidance path for quadrotor under different $\gamma$ coefficients. (b) The boundary distance under different $\gamma$ coefficients. (c) 3D trajectory when $\gamma = 1.3$.}
    \label{fig:vCBF_comparison}
\end{figure}

\begin{figure}[htbp]
    \centering
    \includegraphics[]{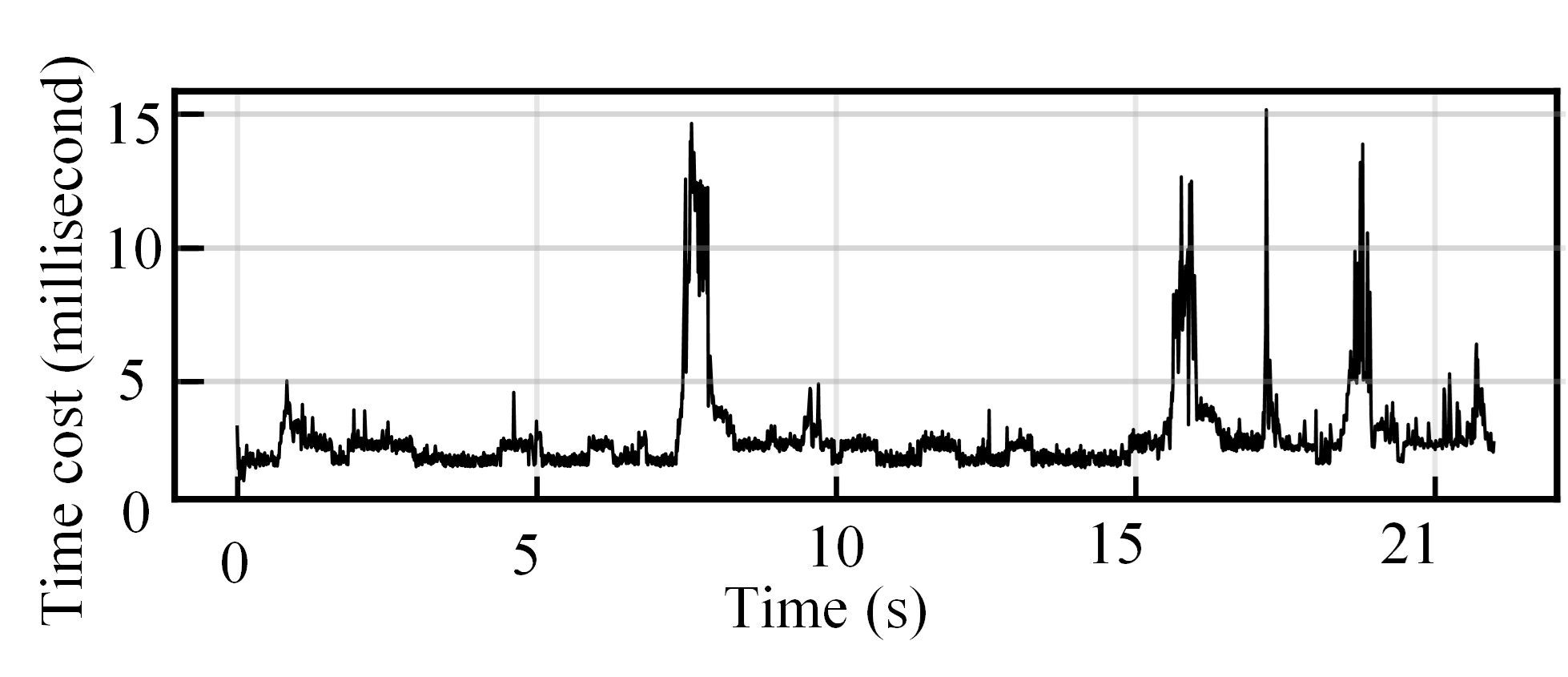}
    \caption{Time cost of optimization problem solving.}
    \label{fig:time_cost}
\end{figure}

The obstacle undergoes elliptical motion with a speed of 2.4 m/s. We verify the performance of obstacle avoidance with different $\gamma$ values. $\gamma = 0$ is equivalent to pure distance-based CBF. Within the MPC time horizon, the trajectory prediction of obstacles is calculated based on a constant speed model. As shown in Fig.~\ref{fig:vCBF_comparison}, a larger $\gamma$ will result in a greater safety margin. From Fig.~\ref{fig:vCBF_comparison}(c), it can be observed that when the quadrotor approaches a high-speed target, it will first slow down, avoid obstacles, and then accelerate. This demonstrates risk management behavior like that of humans. The vCBF constraints can effectively avoid obstacles by considering their dynamic speed, while pure distance-based CBF cannot. This is because the CBF with speed expansion can converge towards the safety set by projecting the relative velocity between the quadrotor and the obstacle onto the unit vector. However, when CBF constraints are added to the standard MPC problem, as illustrated in Fig.~\ref{fig:time_cost}, the computation time increases but remains under 20 milliseconds. However, this is still acceptable. Considering that the MPC of CBF can serve as a path planning for safety constraints, the lower layer relies on MPC to track and achieve real-time safety control of the quadrotor.

\subsubsection{Multi-obstacles path planning}

\begin{figure}[htbp]
    \centering
    \includegraphics[]{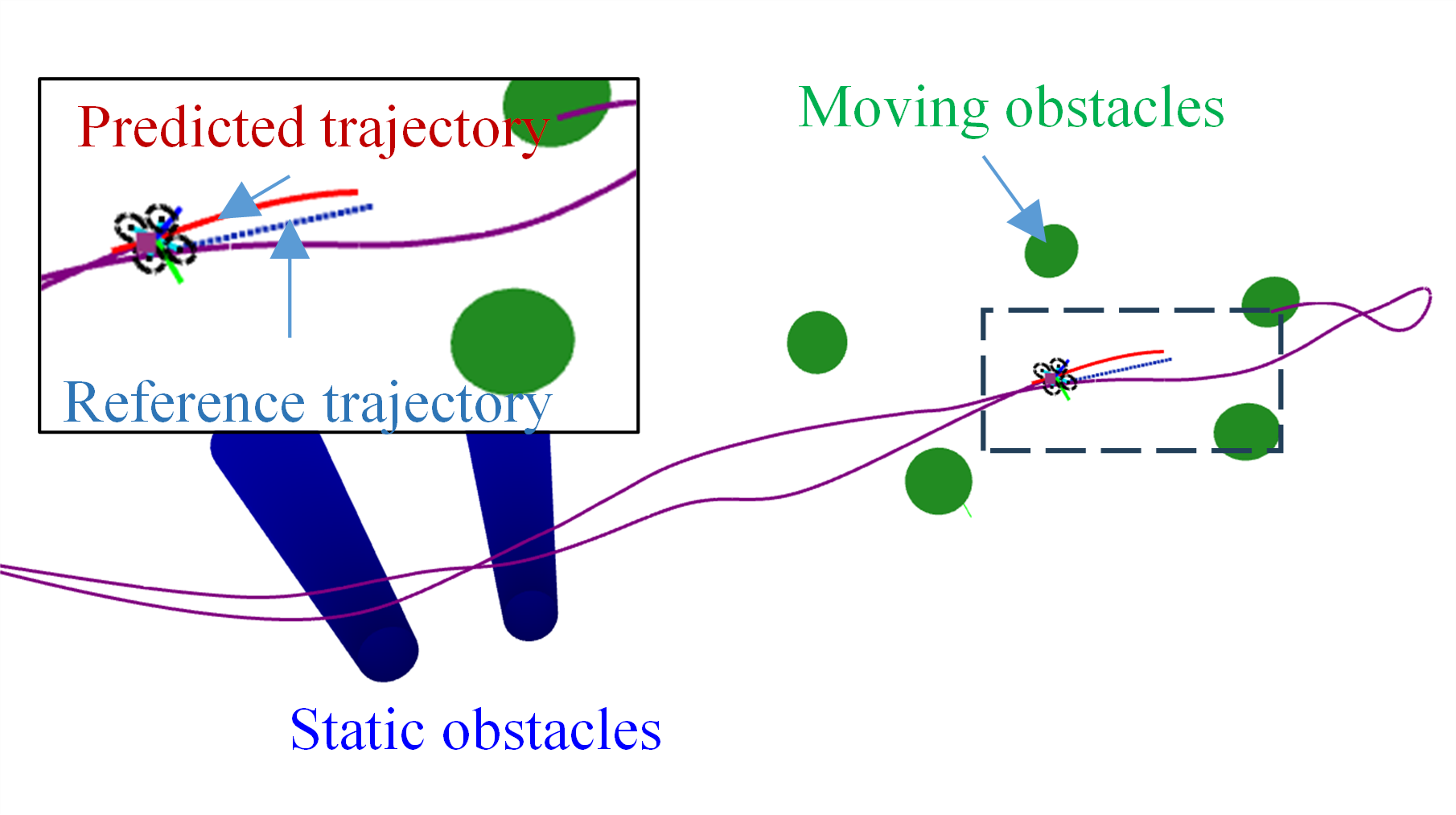}
    \caption{Multi-obstacles trajectory tracking with obstacle avoidance.}
    \label{fig:multi_obstacle}
\end{figure}

As shown in Fig.~\ref{fig:multi_obstacle}, multi-obstacle path planning is also evaluated. The quadrotor successfully navigated the cluttered environment by tracking a dynamically generated, collision-free reference trajectory. The simulator is open source.

\subsection{Experiment platform scheme and results}

\begin{figure}[htbp]
    \centering
        \centering
        \includegraphics[]{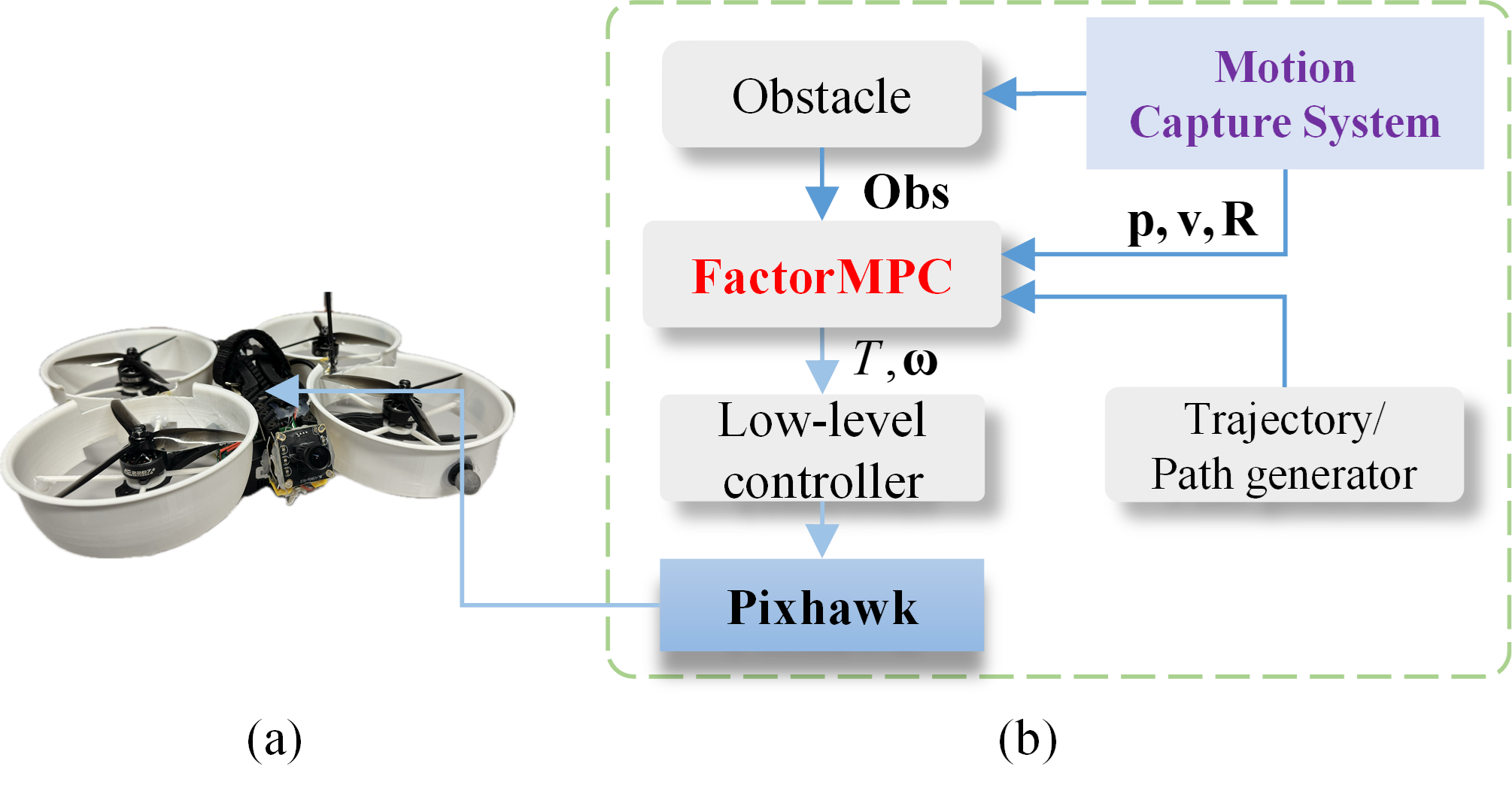}
    \caption{Experimental platform and system framework. (a) The quadrotor is used in experiments. (b) The framework of the quadrotor’s automation system.}
    \label{fig:experiment_platform}
\end{figure}

A self-developed quadrotor is used to conduct trajectory tracking and path planning experiments. The quadrotor and its system framework are illustrated in Fig.~\ref{fig:experiment_platform}(a) and Fig.~\ref{fig:experiment_platform}(b), respectively. The data from the motion capture system (VICON) is used as the ground truth (position, rotation, velocity, and angular velocity). Then, a low-level controller embedded in Pixhawk is adopted to track the desired thrust and angular velocity \cite{ref31}. The proposed method is compared with the differential-flatness-based controller (DFBC).

\subsubsection{Performance comparison between DFBC and FactorMPC}

\begin{figure}[htbp]
    \centering
    \includegraphics[]{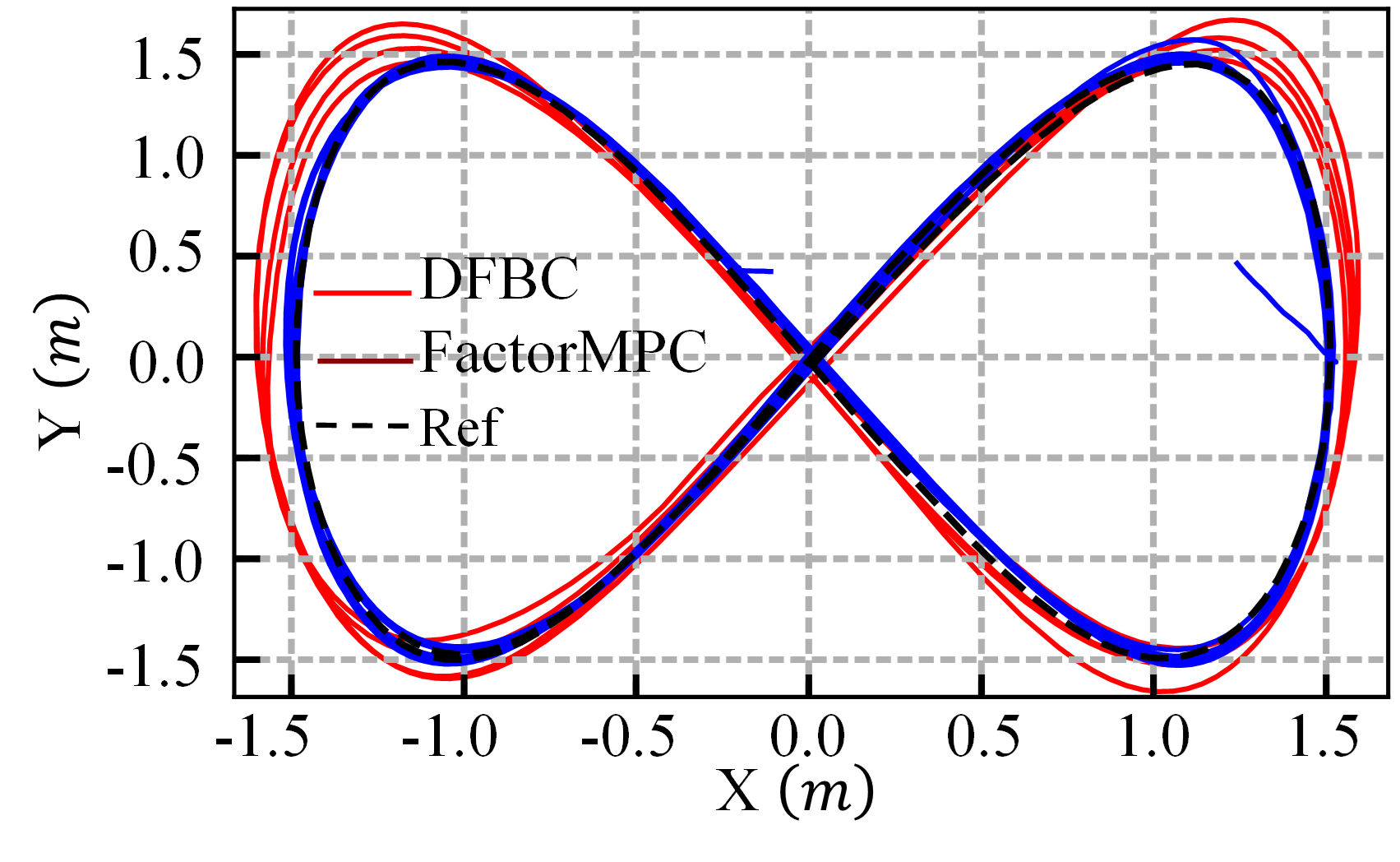}
    \caption{FactorMPC versus DFBC through eight figure tracking tasks.}
    \label{fig:tracking_comparison}
\end{figure}

In experimental evaluations, FactorMPC achieves a control frequency of 100 Hz with an average computation time of 5.4 milliseconds. Notably, 99.6\% of optimizations are completed within 10 milliseconds, demonstrating strong real-time capability. In addition, as shown in Fig.~\ref{fig:tracking_comparison}, the method exhibits smaller tracking errors compared to the baseline DFBC controller.

\subsubsection{FactorMPC demonstrations with obstacle avoidance}

\begin{figure}[htbp]
    \centering
    \includegraphics[]{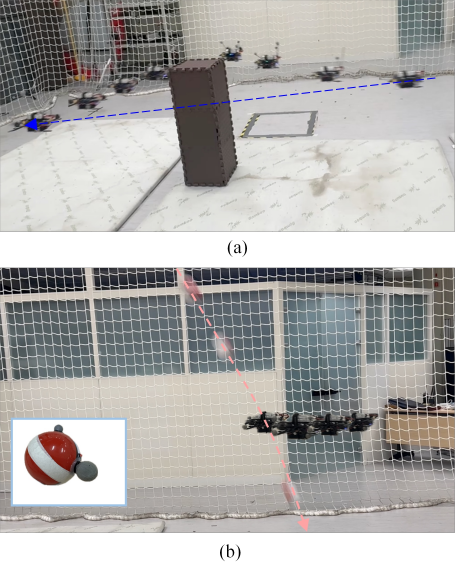}
    \caption{(a) The quadrotor is tracking a straight-line trajectory, blocking with a static obstacle. (b) The quadrotor avoids the rapidly thrown ball and returns to its original position.}
    \label{fig:obstacle_avoidance}
\end{figure}

As shown in Fig. \ref{fig:obstacle_avoidance}, the feasibility of the proposed framework is verified through two experimental demonstrations. In the first scenario, the quadrotor avoids a static obstacle positioned directly in its path. In the second scenario, it detects and avoids a rapidly moving ball (tracked via VICON) before returning to its original position. These experimental results demonstrate that the method effectively generates safer trajectories optimized under CBF constraints.

\section{Conclusion and Future Work}

This paper presented FactorMPC, a novel framework and open-source toolkit that leverages factor graphs to unify planning and control for systems on manifolds. By incorporating geometric properties and defining on-manifold residuals, including innovative vCBF factors for safety, the approach ensures both mathematical fidelity and robust performance. Simulations show the performance of the proposed framework. Experimental validation on quadrotors demonstrated that the toolkit achieves real-time control rates comparable to custom solutions, providing a flexible, modular, and powerful foundation for future advancements in robotics and autonomous systems. Future work will focus on tightly integrating learning by embedding learned dynamics models and neural policies directly as factors within the optimization graph.

 \addtolength{\textheight}{-6cm}   


\bibliographystyle{IEEEtran}
\bibliography{ref}

\end{document}